%% file: main.tex
\documentclass[lettersize,journal]{IEEEtran}
\usepackage{amsmath,amsfonts}
\usepackage{algorithmicx}
\usepackage{algorithm}
\usepackage{algpseudocode}
\usepackage{array}
\usepackage[caption=false,font=scriptsize,labelfont=sf,textfont=sf]{subfig}
\usepackage{textcomp}
\usepackage{stfloats}
\usepackage{url}
\usepackage{verbatim}
\usepackage{graphicx}
\usepackage{cite}
\usepackage[utf8]{inputenc}
\usepackage{ragged2e}
\usepackage{microtype}
\usepackage{pgfplots}
\usepackage{comment}
\usepackage{xcolor}
\usepackage{colortbl}
\definecolor{gray}{rgb}{0.9,0.9,0.9}

\usepackage{adjustbox}
\usepackage{amssymb}
\usepackage{amsmath}
\usepackage{multirow}
\usepackage{capt-of}
\usepackage{makecell}
\usepackage{graphicx}

\newcounter{subfig}

\let\oldtwocolumn\twocolumn
\renewcommand\twocolumn[1][]{%
    \oldtwocolumn[{#1}{%
    \begin{center}
        \refstepcounter{figure} 

        \begin{minipage}{0.3\textwidth}
            \centering
            \includegraphics[width=\linewidth]{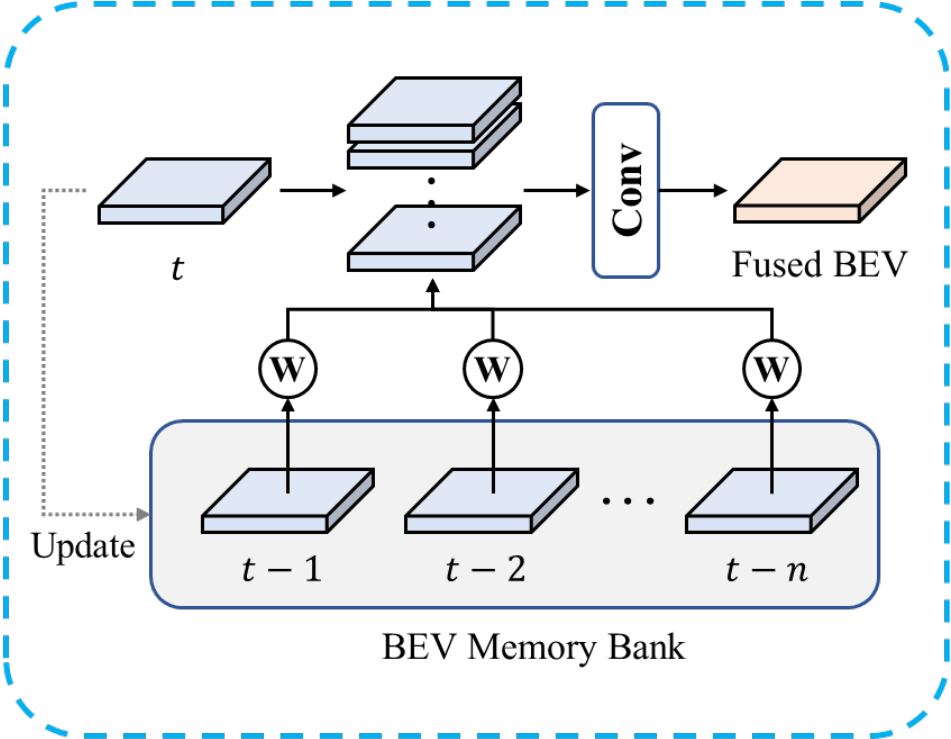}
            \\ (a) Parallel fusion
            \refstepcounter{subfig}
            \label{fig:parallel_fusion}
        \end{minipage}
        \hspace{0.05cm}
        \begin{minipage}{0.3\textwidth}
            \centering
            \includegraphics[width=\linewidth]{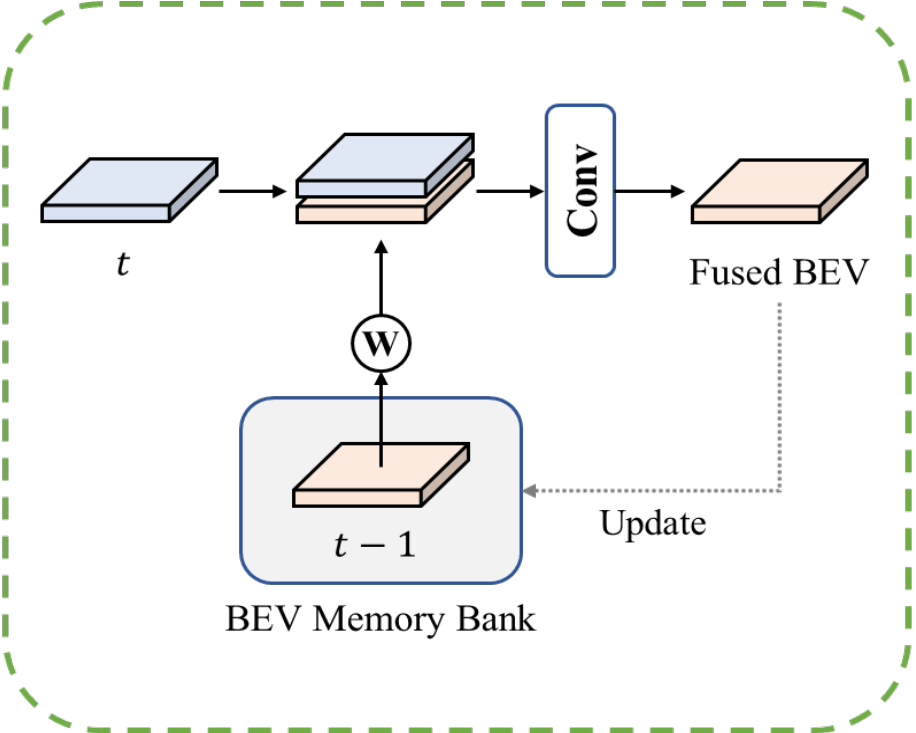}
            \\ (b) Recurrent fusion
            \refstepcounter{subfig}
            \label{fig:recurrent_fusion}
        \end{minipage}
        \hspace{0.05cm}
        \begin{minipage}{0.3\textwidth}
            \centering
            \includegraphics[width=\linewidth]{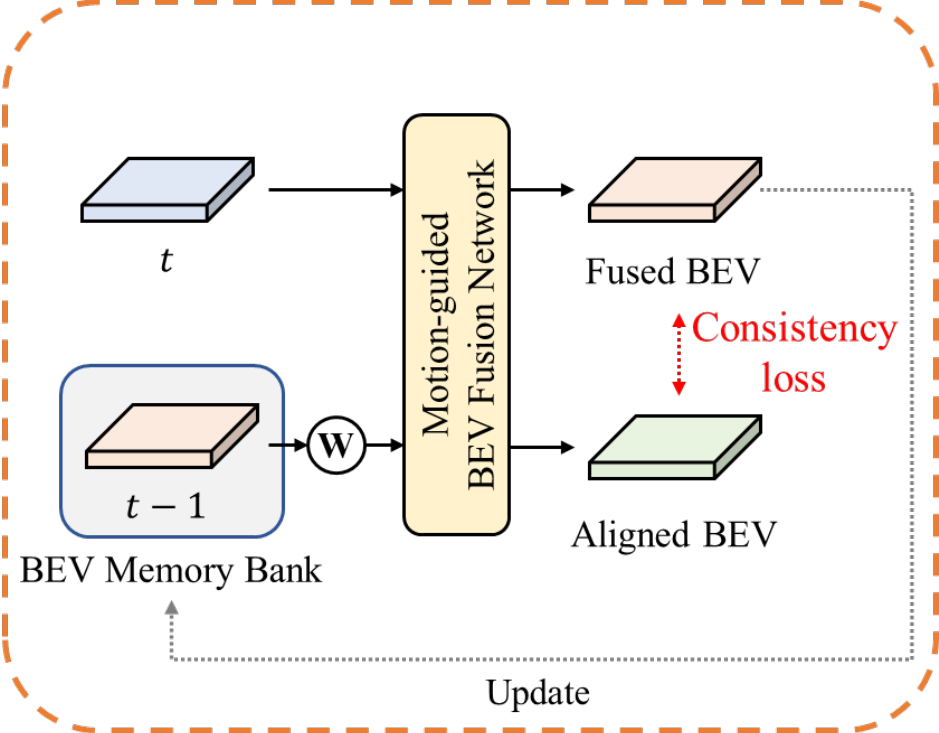}
            \\ (c) Our fusion
            \refstepcounter{subfig}
            \label{fig:our_fusion}
        \end{minipage}
        \addtocounter{figure}{-1}
        \captionof{figure}{\textbf{Different temporal fusion strategies.} 
        (a) Parallel temporal fusion aggregates historical BEV features within a fixed-length window at each time step.
(b) Recurrent temporal fusion progressively updates historical BEV features over time.
(c) Our OnlineBEV approach aligns historical BEV features with current BEV features to enable effective recurrent temporal fusion.}
        \label{fig:motivation_temporal_fusion}
    \end{center}
    }]%
}

\begin{document}

\title{OnlineBEV: Recurrent Temporal Fusion in Bird's Eye View Representations for Multi-Camera 3D Perception}

\author{
    Junho Koh$^{1*}$, Youngwoo Lee$^{2*}$, Jungho Kim$^3$, Dongyoung Lee$^4$, and Jun Won Choi$^4$ \\
    \thanks{$^1$J. Koh is with Autonomous Driving Development Center, Hyundai Motors, 13529 Seongnam, Republic of Korea. (e-mail:  junhkoh@hyundai.com)}
    \thanks{$^2$Y. Lee is with Department of Electrical Engineering,
Hanyang University, 04763 Seoul, Republic of Korea. (e-mail: youngwoolee@spa.hanyang.ac.kr)}
    \thanks{$^3$J. Kim is with Interdisciplinary Program in Artificial Intelligence, Seoul National University, Seoul National University, 08826, Seoul, Republic of Korea (e-mail: jhkim@spa.snu.ac.kr)}
    \thanks{$^4$D. Lee and J. W. Choi are with Department of Electrical and Computer Engineering, Seoul National University, 08826, Seoul,  Republic of Korea. (e-mail: dylee@spa.snu.ac.kr,   junwchoi@snu.ac.kr)}
    \thanks{\textit{(Corresponding author: Jun Won Choi)}}
    \thanks{$^*$denotes equal contribution.}
}
\markboth{Journal of \LaTeX\ Class Files,~Vol.~14, No.~8, July~2025}%
{Shell \MakeLowercase{\textit{et al.}}: A Sample Article Using IEEEtran.cls for IEEE Journals}

\IEEEpubid{0000--0000/00\$00.00~\copyright~2025 IEEE}

\maketitle



\begin{abstract}
Multi-view camera-based 3D perception can be conducted using bird's eye view (BEV) features obtained through perspective view-to-BEV transformations. Several studies have shown that the performance of these 3D perception methods can be further enhanced by combining sequential BEV features obtained from multiple camera frames. However, even after compensating for the ego-motion of an autonomous agent, the performance gain from temporal aggregation is limited when combining a large number of image frames. This limitation arises due to dynamic changes in BEV features over time caused by object motion. In this paper, we introduce a novel temporal 3D perception method called OnlineBEV, which combines BEV features over time using a recurrent structure. This structure increases the effective number of combined features with minimal memory usage. However, it is critical to spatially align the features over time to maintain strong performance. OnlineBEV employs the Motion-guided BEV Fusion Network (MBFNet) to achieve temporal feature alignment. MBFNet extracts motion features from consecutive BEV frames and dynamically aligns historical BEV features with current ones using these motion features. To enforce temporal feature alignment explicitly, we use Temporal Consistency Learning Loss, which captures discrepancies between historical and target BEV features. Experiments conducted on the nuScenes benchmark demonstrate that OnlineBEV achieves significant performance gains over the current best method, SOLOFusion. OnlineBEV achieves 63.9\% NDS on the nuScenes test set, recording state-of-the-art performance in the camera-only 3D object detection task.

\end{abstract}



\section{Introduction}
\IEEEPARstart{3}{D} perception plays a crucial role in autonomous driving and robot navigation by gathering comprehensive information about the surrounding 3D environment through sensor data. It encompasses various tasks such as 3D object detection, bird's eye view (BEV) segmentation, and 3D occupancy grid prediction. 

\IEEEpubidadjcol

Multi-view cameras covering a 360-degree perspective have enabled the identification of a 3D environment around the ego-vehicle.
Recent research efforts have been dedicated to a task of transforming multiple 2D images into a unified 3D representation for 3D perception.
Existing architectures for generating 3D representations from multi-view camera images can be categorized into two strategies: dense BEV-based methods \cite{bevdet, bevdepth, aedet, fbbev, sabev} and sparse query-based methods \cite{detr3d, petr, vedet}.
BEV-based methods utilize the lift-Splat-Shoot (LSS) mechanism \cite{lss} to convert features extracted from 2D images into unified representations within the BEV domain. On the other hand, the query-based methods leverage attention mechanisms to decode object query features by utilizing multi-view 2D features.

Baseline models for 3D perception process single-frame images captured by multi-view cameras at each time step. This approach may underperform when the current frame experiences occlusion or motion blur. To compensate for this degradation, it is beneficial to use multiple consecutive frames for 3D perception through a technique known as temporal fusion. In this technique, features from historical frames are utilized to enhance robustness in perception tasks.

There are two main strategies for temporal fusion: (1) parallel temporal fusion and (2) recurrent temporal fusion, as illustrated in Fig. \ref{fig:motivation_temporal_fusion} (a) and Fig. \ref{fig:motivation_temporal_fusion} (b). In parallel temporal fusion, information from the most recent $K$ frames is aggregated simultaneously, requiring the memory bank that stores features from all $K$ frames. Furthermore, the computational complexity of this approach increases with larger $K$, which limits the number of frames that can be combined. In contrast, recurrent temporal fusion maintains and updates a single feature that encapsulates historical information in a recurrent manner. This design supports long-term temporal fusion while keeping the computational complexity low.

\begin{figure*}[t]
	\centering
        \centerline{\includegraphics[width=0.99\textwidth]{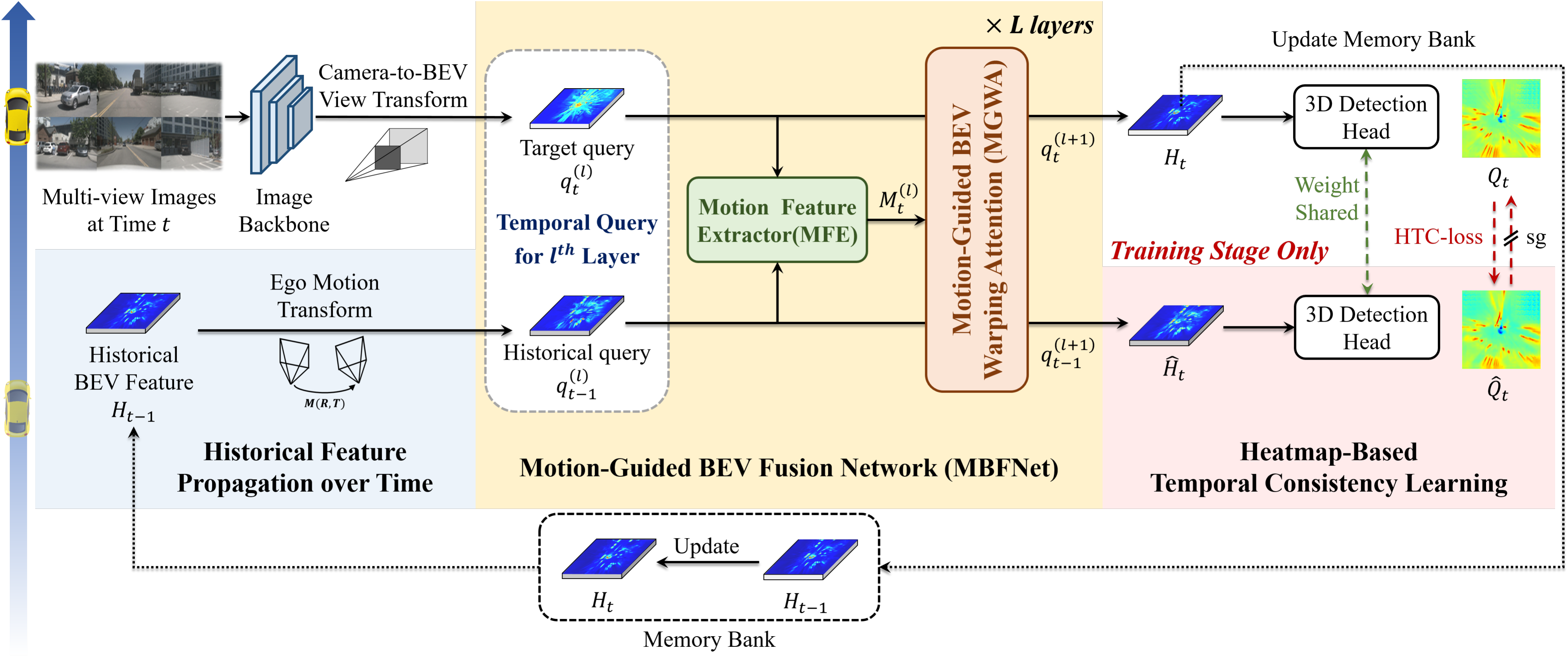}}
        \caption {\textbf{The overall architecture of OnlineBEV.} OnlineBEV aggregates historical BEV features with current BEV features using a recurrent structure. Before aggregation, MGWA aligns the historical BEV features to the current ones, guided by the motion features produced by MFE. During training, HTC-loss further facilitates the feature alignment process. Here, `sg' denotes stop-gradient.}
	\label{fig:overall}
\end{figure*}

Parallel temporal fusion has been employed in 3D perception frameworks such as \cite{bevdet4d, bevdepth, bevstereo, p2d, solofusion, sparsebev}. Notably, SOLOFusion achieved significant computational complexity reduction by storing previous BEV features in memory rather than the raw sensor data. Recurrent temporal fusion has also been adopted in StreamPETR \cite{streampetr}, which integrates features from historical frames using a recurrent architecture. While StreamPETR achieved high computational efficiency through its sparse, object-centric design, it lacks a unified BEV representation. This limitation constrains its applicability to broader 3D scene understanding tasks—such as BEV segmentation and 3D occupancy prediction—that require dense, spatially aligned features. In this paper, we aim to design recurrent temporal fusion architecture tailored for BEV representations. 




One of the key challenges in temporal fusion lies in handling significant spatial changes within dynamic scenes, which can cause substantial feature misalignments across frames. If not properly addressed during feature aggregation, these misalignments may propagate over time, leading to a noticeable degradation in performance. Moreover, spatial changes induced by object motion are often more pronounced in the BEV domain than in the perspective view. 
Therefore, it is essential to both align and appropriately weight features from different frames to fully leverage the advantages of temporal fusion.

In this paper, we propose a novel recurrent temporal fusion framework, OnlineBEV, for multi-camera 3D perception. The objective of OnlineBEV is to effectively aggregate BEV features across frames using a recurrent design, as illustrated in Fig. \ref{fig:motivation_temporal_fusion} (c). While enabling long-term temporal fusion, OnlineBEV maintains high performance by compensating for spatial misalignment over time through the Motion-Guided BEV Fusion Network (MBFNet).

OnlineBEV introduces two key innovations. First, it utilizes MBFNet to align BEV feature maps across time using a spatio-temporal deformable attention mechanism. This alignment is guided by two core modules: the Motion Feature Extractor (MFE) and the Motion-Guided BEV Warping Attention (MGWA). MFE captures spatial changes between adjacent BEV features and encodes them into motion features. MGWA then leverages deformable attention \cite{def-detr} to spatially align the features based on these motion cues.
Second, we introduce a novel temporal consistency learning strategy that explicitly guides the alignment of BEV features. By enforcing consistency between the heatmap representations of historical and current BEV feature maps, the proposed consistency loss significantly improves the temporal alignment capability of MBFNet. Experimental results confirm that this approach enhances both the accuracy and robustness of feature fusion over time.

The proposed OnlineBEV framework is evaluated on the public nuScenes benchmark \cite{nuscenes}. Our evaluation shows that OnlineBEV achieves significant performance gains over baseline models across all tasks, including 3D object detection, BEV segmentation, and 3D occupancy prediction. Additionally, OnlineBEV outperforms other existing methods in the official nuScenes 3D object detection benchmark.

The key contributions of our work are summarized as follows:
\begin{itemize}
\item We propose OnlineBEV, a novel temporal fusion framework that efficiently aligns and aggregates sequential BEV features for multi-view 3D perception.
\item OnlineBEV leverages the advantages of the recurrent fusion strategy while achieving strong performance through a robust feature alignment mechanism enabled by the proposed MBFNet.
\item We present MBFNet that spatially aligns BEV features by utilizing motion features capturing the spatial changes between previous and current BEV feature maps.
\item We further introduce a temporal consistency learning strategy that minimizes the difference between historical and current BEV features, thereby enhancing alignment quality and temporal coherence.
\end{itemize}

\section{Related Work}
\subsection{Multi-view 3D Perception}


Multi-view 3D perception has rapidly advanced by transforming feature maps from multiple viewpoints into a unified 3D representation, enabling more efficient processing of multi-camera inputs compared to single-view perception~\cite{haq2022one, yao2023occlusion, chen2023shape}. Based on their transformation strategies, multi-view 3D perception frameworks can be broadly categorized into dense BEV-based methods~\cite{bevdet, bevdepth, aedet, fbbev, sabev} and sparse query-based methods~\cite{detr3d, petr}.

Dense BEV-based approaches adopt the LSS paradigm~\cite{lss} to project 2D perspective image feature maps into a unified BEV representation. In contrast, sparse query-based methods~\cite{detr3d, petr, vedet} employ learnable object queries and 3D position-aware features to directly aggregate multi-view image features using attention mechanisms. Despite their effectiveness, these frameworks are limited by their reliance on single-frame inputs, lacking temporal context from image sequences.

Recently, both paradigms have been extended to incorporate temporal modeling. BEVFormer~\cite{bevformer} pioneered temporal modeling for multi-view 3D object detection by introducing a temporal attention mechanism. Dense BEV-based methods have since achieved significant performance gains by fusing adjacent BEV feature maps across time. For instance, BEVStereo~\cite{bevstereo} improved depth estimation by applying multi-view stereo (MVS)\cite{mvs} to consecutive keyframes, while SOLOFusion\cite{solofusion} enhanced long-term temporal fusion by storing historical BEV features in a memory bank and integrating them with current features through lightweight processing.

Sparse query-based approaches have similarly evolved to exploit temporal information. DETR4D~\cite{detr4d} and Sparse4D~\cite{sparse4d} introduced temporal attention to enable interactions between past and current object queries. StreamPETR~\cite{streampetr} propagated long-term historical object proposals as queries, encoding ego-motion and surrounding object dynamics for temporal reasoning. SparseBEV~\cite{sparsebev} defined object queries in BEV space and utilized adaptive spatio-temporal sampling and mixing to interact with multi-view image features for 3D object prediction.

\begin{figure*}[t]
	\centering
        \centerline{\includegraphics[width=0.75\textwidth]{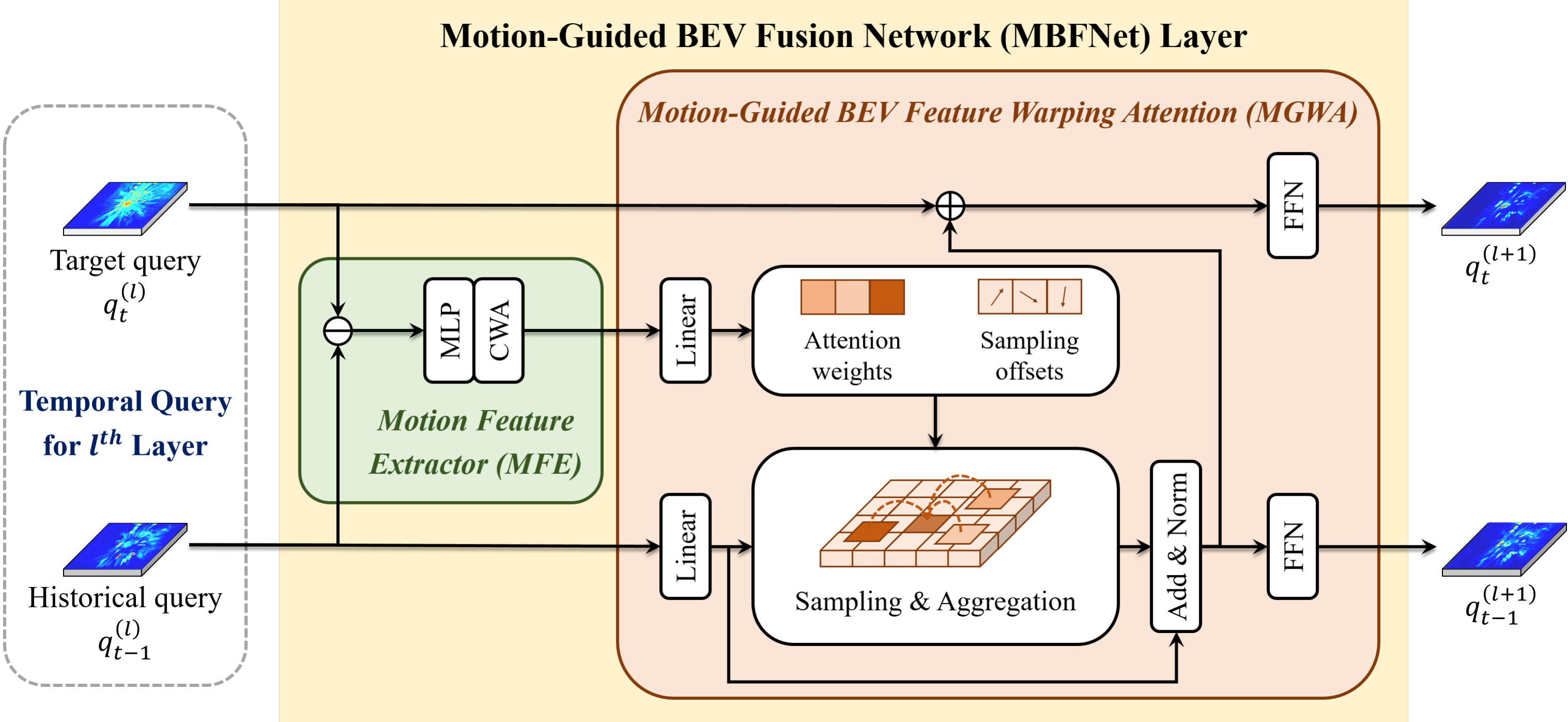}}
        \caption {\textbf{Structure of Motion-Guided BEV Fusion Network (MBFNet).} MBFNet comprises MFE and MGWA, where MFE generates motion features and MGWA applies deformable attention for feature alignment and fusion.  }
	\label{fig:mbfnet}
\end{figure*}

\subsection{Consistency Learning}
Consistency learning has been widely explored as a strategy to enhance feature representations by leveraging both labeled and unlabeled datasets. For example, CSD~\cite{csd} introduced a consistency loss by enforcing constraints between foreground proposals predicted from the original input image and its horizontally flipped counterpart. Similarly, CCT~\cite{cct} proposed a robust and efficient consistency learning approach that exploits the invariance of predictions to various perturbations at the feature level.

Recently, consistency learning has been extended to videos to exploit temporal correspondences across frames. To stabilize predictions for video inputs, the method in~\cite{tc} utilized temporal consistency by incorporating optical flow between consecutive frames. Furthermore, consistency learning has also been applied to exploit temporal information without relying exclusively on unsupervised datasets. MaskFreeVIS~\cite{maskfreevis}, for instance, applied a temporal consistency loss between one-to-many matched patches across consecutive frames, improving instance segmentation by capturing complex object shapes through temporal information.

Building on these advances, we propose a novel temporal consistency learning strategy designed specifically for feature alignment. This approach guides the alignment of features from the previous to the current time step, thereby enhancing 3D perception performance.


\section{OnlineBEV}
The overall architecture of the proposed OnlineBEV is illustrated in Fig. \ref{fig:overall}. An image backbone network with shared weights processes multi-camera images at time $t$ to produce multi-view feature maps. To generate the BEV feature $F_t$, we adopt the LSS method \cite{lss}, which efficiently transforms perspective features into the BEV space using depth predictions. Notably, our framework can also incorporate alternative BEV generation methods, such as inverse perspective mapping (IPM) or BEVFormer \cite{bevformer}.
OnlineBEV maintains historical features $H_{t-1}$ in memory, which are updated in a recurrent manner. To conduct temporal fusion, we devise a target query $q_t^{(l)}$ and a historical query $q_{t-1}^{(l)}$, where $l$ denotes the transformer layer. The target query is initialized with $F_t$ while the historical query is initialized with $H_{t-1}$. 

In each transformer layer, MGWA aligns the historical query $q_{t-1}^{(l)}$ with the target query $q_t^{(l)}$, guided by the motion features produced by MFE. To generate these motion features, MFE leverages both the target and historical queries. The aligned historical query $\hat{q}_{t-1}^{(l)}$ is then updated as the historical query $q_{t-1}^{(l+1)}$ for the subsequent layer. Simultaneously, the target query $q_t^{(l+1)}$ for the next layer is obtained by fusing $\hat{q}_{t-1}^{(l)}$ with $q_t^{(l)}$. Through this iterative process, the historical query progressively converges toward the target query, ultimately producing enhanced BEV features.  

After passing through $L$ transformer layers, the queries are transformed into the final target query $q_t^{(L)}$ and the final historical query $q_{t-1}^{(L)}$. We let $H_t = q_t^{(L)}$ and $\hat{H}_t = q_{t-1}^{(L)}$, where $\hat{H}_t$ represents the features temporally aligned with $H_t$. To further enhance alignment, we introduce the Heatmap-based Temporal Consistency Loss (HTC-loss), which encourages greater similarity between the heatmaps derived from $H_t$ and $\hat{H}_t$. Finally, a detection head is applied to  $H_t$  to produce the final output and  $H_t$ is stored in the memory bank for use in the next time step.

\subsection{Motion-Guided BEV Fusion Network}
Fig. \ref{fig:mbfnet} presents the structure of the MBFNet.
MBFNet consists of MFE and MGWA. MGWA spatially aligns the historical BEV features with the current BEV features using motion context information produced by MFE. Then, MGWA aggregates the aligned features with the current BEV features to yield the enhanced BEV representation. 

\subsubsection{Motion Feature Extractor}
MFE performs differential encoding between the historical query \( q_{t-1}^{(l)} \) and the target query \( q_{t}^{(l)} \) to extract motion features. It generates a motion map by computing the difference between \( q_{t-1}^{(l)} \) and \( q_{t}^{(l)} \). Static objects are expected to exhibit minimal changes across frames, whereas dynamic objects produce significant feature differences over time. By encoding these variations, the network effectively captures object motion within the BEV space. The resulting difference is further processed through fully connected (FC) layers followed by Channel-Wise Attention (CWA) \cite{senet}, yielding the motion context feature \( M_t^{(l)} \),
 i.e.,
\begin{align}
     M_t^{(l)} &= \mathrm{CWA}(\mathrm{FC}(q_{t}^{(l)}-q_{t-1}^{(l)})).
     \label{eq: motion}
\end{align}

\subsubsection{Motion-Guided BEV Warping Attention}
Using the motion features $M_t^{(l)}$  extracted by MFE, MGWA aligns the historical query $q_{t-1}^{(l)}$ with the target query $q_{t}^{(l)}$. The feature alignment is achieved by a deformable attention mechanism originally proposed in \cite{def-detr}. The motion features $M_{t}^{(l)}$ is used to determine the offsets and attention weights of a mask, which is applied to  $q_{t-1}^{(l)}$. 
At the reference point $p$, the aligned features $Z^{(l)}_{t-1}(p)$ are given by 
\begin{align}
    &Z^{(l)}_{t-1}(p)=\mathrm{DeformAttn}(M_{t-1}^{(l)}(p), p, q_{t-1}^{(l)}),
\end{align}
where $\mathrm{DeformAttn}(Q,p,V)$ is Deformable Cross Attention module \cite{def-detr}, and $Q$ and $V$ denote query and value, respectively. 
The historical query $q^{(l+1)}_{t-1}$ for the next iteration is finally updated as
\begin{gather}
    q^{(l+1)}_{t-1}= \mathrm{FFN_{1}}(\hat{q}^{(l)}_{t-1}), \\
    \hat{q}^{(l)}_{t-1} = \mathrm{LN}(\mathrm{Dropout}(Z^{(l)}_{t-1}) + q_{t-1}^{(l)}), 
\end{gather}
where $\mathrm{FFN_{1}}(\cdot)$ is the feed-forward network \cite{def-detr}, $\mathrm{Dropout}(\cdot)$ is the Dropout operation \cite{dropout}, and $\mathrm{LN}(\cdot)$ is the Layer Normalization \cite{layernorm}. 
The target query $q_{t}^{(l+1)}$ for the next iteration is also updated by combining the aligned historical query $\hat{q}^{(l)}_{t-1}$ with the target query $q_{t}^{(l)}$, i.e., 
\begin{align}
    q_{t}^{(l+1)} = \mathrm{FFN_{2}}(q_{t}^{(l)} \oplus \hat{q}^{(l)}_{t-1}),
\end{align}
where $\oplus$ denotes channel-wise concatenation and $\mathrm{FFN_{2}}(\cdot)$ is the feed-forward network composed of one fully-connected layer with ReLU activation. 

After $L$ transformer layers, the final aggregated BEV features $q_{t}^{(L)}$ and the aligned historical BEV features $q_{t-1}^{(L)}$ become $H_{t}$ and $\hat{H}_{t}$, respectively.
The final object detection results are obtained by applying the 3D detection head to $H_t$ \cite{centerpoint}.
Note that $H_{t}$ is returned to the memory queue for BEV detection in the subsequent time step.

\subsection{Heatmap-Based Temporal Consistency Loss} 
HTC-loss provides explicit supervision for feature alignment during training by enforcing consistency between the current BEV feature $H_t$ and the temporally aligned historical BEV feature $\hat{H}_t$. This promotes stability and coherence in BEV representations across consecutive frames, enabling more effective long-term temporal fusion.

To compute HTC-loss, two prediction heads with shared weights~\cite{centerpoint} are applied to $H_t$ and $\hat{H}_t$, producing heatmaps $Q_t$ and $\hat{Q}_t$, respectively. The HTC-loss is then defined as
\begin{equation}
    \mathcal{L}_{\text{cons}} = \lVert Q_t - \hat{Q}_t \rVert_2^2.
\end{equation}
Here, gradient flow is blocked for the branch generating $Q_t$ to ensure that $\hat{H}_t$ is aligned to the target BEV features $H_t$.

\subsection{Overall Training Loss Function}
The proposed OnlineBEV is an end-to-end trainable with a total loss 
\begin{equation}
    \mathcal{L}=\omega_{cls} \mathcal{L}_{cls} + \omega_{reg} \mathcal{L}_{reg} + \omega_{cons} \mathcal{L}_{cons},
\end{equation}
where $w_{cls}$, $w_{reg}$, and $w_{cons}$ are the regularization parameters, and $\mathcal{L}_{cls}$, $\mathcal{L}_{reg}$, and $\mathcal{L}_{cons}$ are  the focal loss \cite{focalloss}, the L1 loss for 3D box regression, and the HTC-loss, respectively.
 We set $w_{cls}$ and $w_{reg}$ to 1 and 0.25, respectively, following the BEVDepth \cite{bevdepth}. 
Meanwhile, $w_{cons}$ is set to 2 based on our experiments. More details of training procedures are provided in Section \ref{section:implementation_details}.

\input{table/sota_val}
\input{table/sota_test}
\input{table/perception_ablation}

\section{Experiments}
\subsection{Datasets and Performance Metrics}
The nuScenes dataset \cite{nuscenes} is a challenging large-scale autonomous driving dataset comprising 1,000 scenes, each with a duration of about 20 seconds. These 1000 driving scenes are split into 700 scenes for training (\textit{train}), 150 validation (\textit{val}), and 150 for testing (\textit{test}). Six cameras provide perspective-view images, covering the entire $360^{\circ}$ field of view (FOV). Moreover, 3D bounding boxes from 10 categories are annotated at 2Hz. The proposed method was evaluated in regarding mean average precision (mAP) and nuScenes detection score (NDS), which are official 3D object detection benchmark metrics of nuScenes. mAP is computed based on the 2D center distance between the ground truth data and the predictions at the BEV. NDS is a weighted sum of mAP and 5 kinds of true positive (TP) metrics including average translation error (mATE), average scale error (mASE), average orientation error (mAOE), average velocity error (mAVE), and average attribute error (mAAE).

The Argoverse 2 dataset~\cite{argoverse2} comprises 1,000 driving scenes, each lasting 15 seconds, with annotations provided at 10\,Hz. The dataset is divided into 700 scenes for training, 150 for validation, and 150 for testing. It features seven high-resolution ring cameras providing a complete $360^\circ$ field of view. Annotations cover 26 object categories within a sensing range of 150\,meters. 
We evaluate our method using the official Argoverse 2 3D detection metrics. In addition to mAP, the primary evaluation metric is the Composite Detection Score (CDS), defined as a weighted sum of mAP and three TP metrics: mATE, mASE, and mAOE.



\subsection{Implementation Details} \label{section:implementation_details}

We conducted experiments on the nuScenes dataset. We employed ResNet50, ResNet101~\cite{resnet}, and V2-99~\cite{vovnet} as image backbones. When we employed ResNet50, the input resolution of multi-view images was set to $256 \times 704$, and the size of BEV feature was set to $128 \times 128$. For larger backbones such as ResNet101 and V2-99, the input image sizes were set to $512 \times 1408$ and $640 \times 1600$, respectively, with the BEV resolution configured to $256 \times 256$.

OnlineBEV was trained in two phases. In the first phase, a single-frame model without MBFNet and HTC-loss was trained for 6 epochs. In the second phase, the full OnlineBEV model was trained with the recurrent structure  for a total of 90 epochs using the AdamW optimizer~\cite{adamw} with a learning rate of 2e-4. We used three transformer layers with a dropout rate of 0.1 during training, and dropout was disabled during inference. A batch size of 32 was used with the ResNet50 backbone on 4 NVIDIA RTX 3090 GPUs. For larger backbones (ResNet101 and V2-99), the batch size was reduced to 16, and training was conducted on 8 NVIDIA Tesla V100 GPUs. We adopted data augmentation used in \cite{bevdepth}.

To train the recurrent structure, we utilized a sequential dataloader that preserves the temporal order of frames within each sequence. Frames were processed sequentially, and all historical frames were ego-motion compensated prior to temporal fusion.  To avoid feature contamination across scenes, the memory was reset to zero at the start of each new scene. OnlineBEV and other competitive methods were trained and evaluated without access to future frames.

When we trained the OnlineBEV on the Argoverse 2 dataset,  we used the V2-99 backbone with an input resolution of $640 \times 960$ and a BEV resolution of $256 \times 256$. We used the same optimization settings and training strategies as used in the nuScenes experiments. The model was trained for 6 epochs with a batch size of 8 on 8 NVIDIA Tesla V100 GPUs.

\input{table/main_ablation}
\input{table/alignment_ablation}

\subsection{Performance Comparison}
\subsubsection{Performance on nuScenes Valid Set}

Table~\ref{tab:sota_val} summarizes the performance of \textit{OnlineBEV} compared to existing multi-view 3D object detectors on the nuScenes \textit{validation} set. Here, \# Frames denotes the number of frames used during training, $\dagger$ indicates methods leveraging perspective-view pre-training~\cite{dd3d}, and \textit{rnt} denotes a recurrent structure.  
\textit{OnlineBEV} achieves substantial performance gains over other temporal fusion approaches. With a ResNet-50 backbone pretrained on ImageNet-1k~\cite{imagenet}, \textit{OnlineBEV} delivers a 1.7\% improvement in mAP and a 1.1\% increase in NDS compared to SOLOFusion~\cite{solofusion}, which uses 16 historical frames. In contrast, \textit{OnlineBEV} maintains only a single historical feature map.  
When ResNet-50 is pretrained on nuImages~\cite{nuscenes}, \textit{OnlineBEV} achieves state-of-the-art results with 46.3\% mAP and 56.0\% NDS. Furthermore, with a ResNet-101 backbone and an input resolution of $512 \times 1408$, \textit{OnlineBEV} surpasses the previous best, StreamPETR~\cite{streampetr}, by 0.4\% in mAP and 0.7\% in NDS.

\subsubsection{Performance on nuScenes Test Set}

Table \ref{tab:sota_test} provides the performance of OnlineBEV evaluated on the nuScenes test set. 
 Note that ConvNeXt-B~\cite{convnext} backbone is pretrained on ImageNet-22K \cite{imagenet}, and V2-99 \cite{vovnet} backbone is initialized with weights from DD3D \cite{dd3d}.
Using the V2-99 \cite{vovnet} backbone, OnlineBEV outperforms other temporal fusion methods. The performance of OnlineBEV surpasses that of the latest state-of-the-art method, SparseBEV \cite{sparsebev}, by 0.2\% in mAP and 0.3\% in NDS.  Additionally, our OnlineBEV outperforms SOLOFusion \cite{solofusion} configured with 16 historical frames, by 1.8\% in mAP and 2.0\% in NDS.



\subsubsection{Performance on Other 3D Perception Tasks} \label{experiments:3d_perception}
Table~\ref{table:perception_ablation} demonstrates the effectiveness of \textit{OnlineBEV} on additional 3D perception tasks, including BEV segmentation and 3D occupancy prediction. Although numerous sophisticated methods exist, we compare \textit{OnlineBEV} with representative temporal fusion approaches, BEVDepth~\cite{bevdepth} and SOLOFusion~\cite{solofusion}.  
For the BEV segmentation task, \textit{OnlineBEV} achieves a 7.7\% mean Intersection-over-Union (mIoU) improvement over BEVDepth and a 2.9\% gain over SOLOFusion. In the 3D occupancy grid prediction task, \textit{OnlineBEV} outperforms BEVDepth by 3.3\% mIoU and SOLOFusion by 1.2\%.

\input{table/agoverse}
\input{table/data_corruption_ablation}

\subsection{Ablation Studies}
We performed a series of ablation studies on the nuScenes \textit{validation} set to evaluate the contributions of each sub-module. Two baseline models were considered: Baseline-S, which processes a single frame without temporal fusion, and Baseline-M, which fuses 2, 5, and 17 frames using a parallel temporal fusion strategy. For clarity, we denote these configurations as Baseline-M (2), Baseline-M (5), and Baseline-M (17).  
Furthermore, we applied MGWA to Baseline-M (2) and Baseline-M (5). However, applying MGWA to Baseline-M (17) was not feasible due to memory limitations.

\begin{figure*}[t]
    \centering
    \subfloat[]{\includegraphics[width=0.27\textwidth]{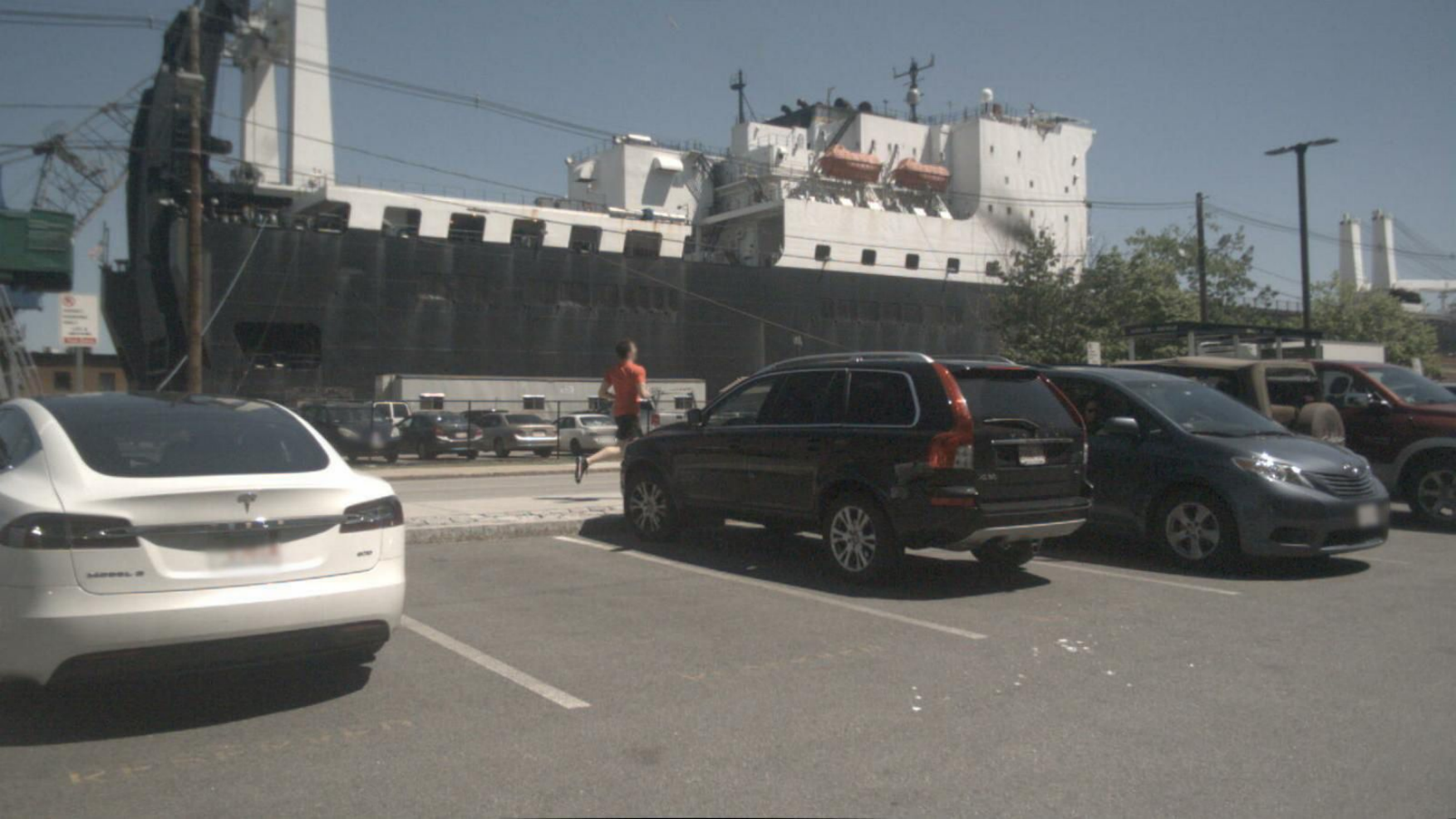}}
    \hspace{0.2cm}
    \subfloat[]{\includegraphics[width=0.27\textwidth]{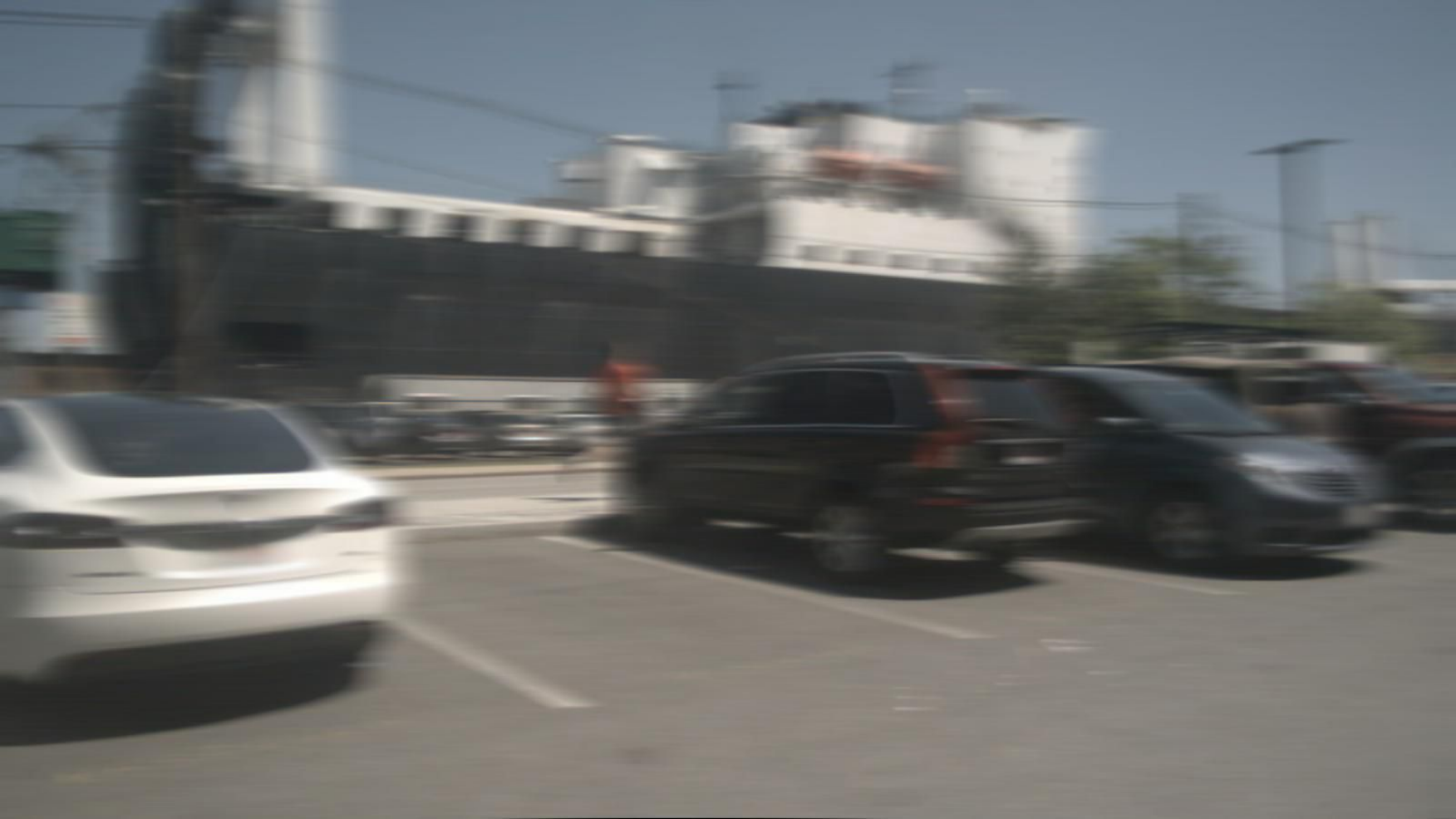}}
    \hspace{0.2cm}
    \subfloat[]{\includegraphics[width=0.27\textwidth]{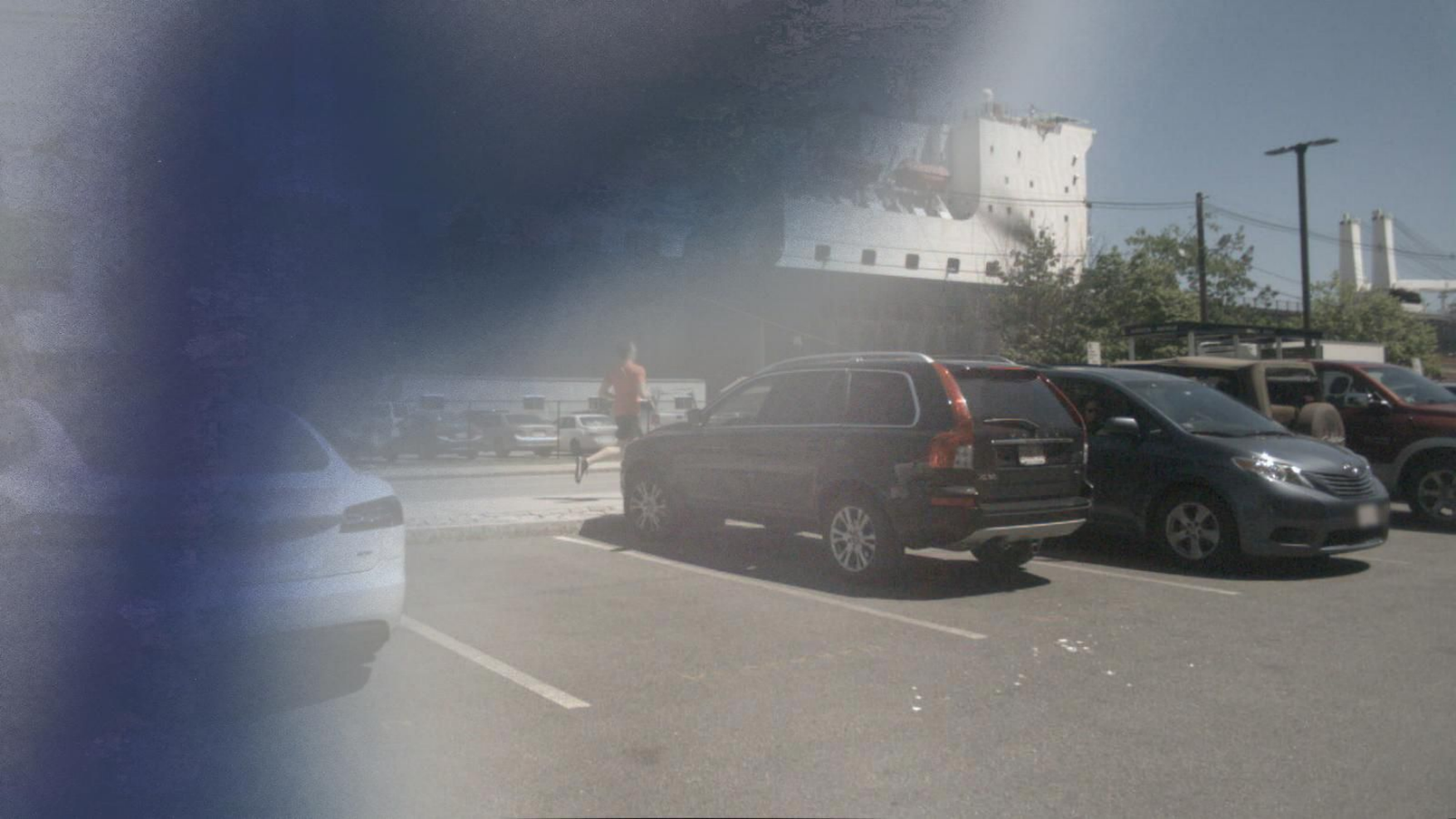}}

    \caption{\textbf{Examples of corrupted input images on nuScenes dataset.} 
    (a), (b), and (c) show the original input, motion blur, and occlusion, respectively.}
    \label{fig:data_corruption}
\end{figure*}

\begin{figure*}[t]
	\centering
        \centerline{\includegraphics[width=0.85\textwidth]{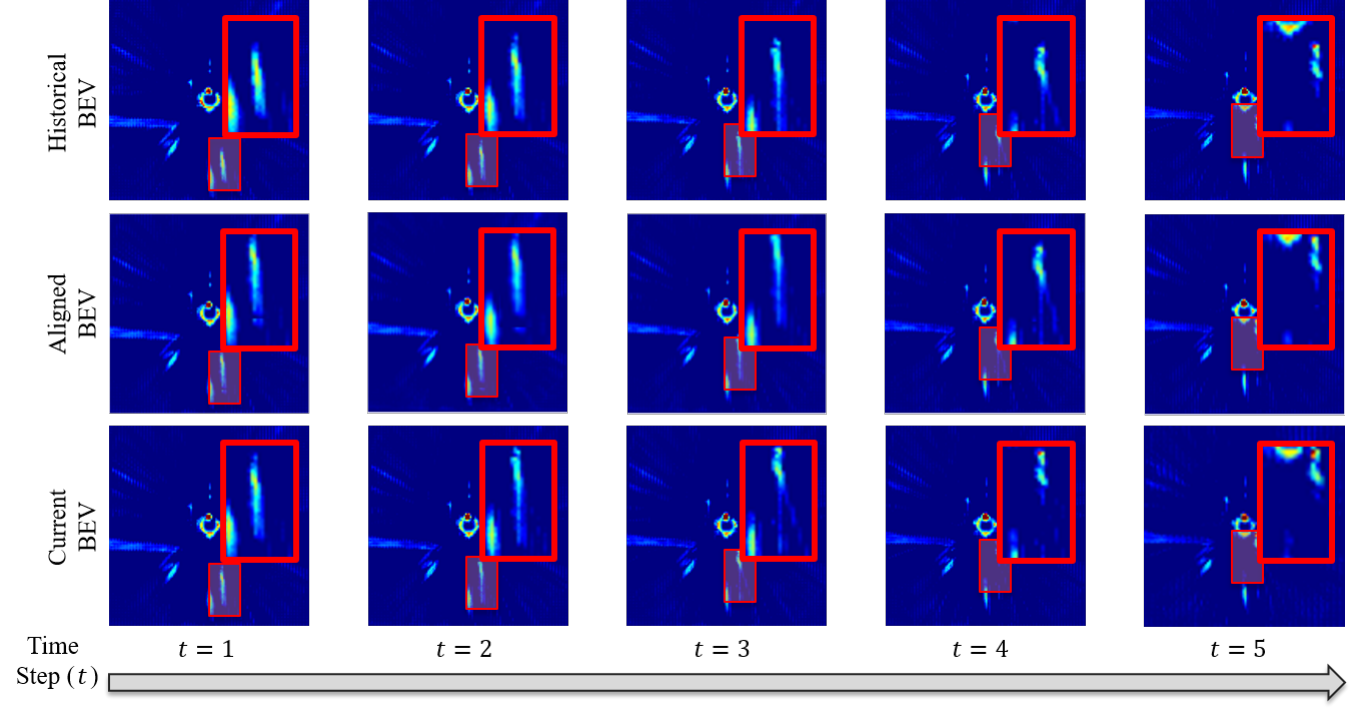}}
        \caption{\textbf{Visualization of BEV features over time.} Historical, aligned, and current BEV features are visualized over five consecutive time steps ($t=1$ to $t=5$) to show the temporal evolution and the effectiveness of the alignment process.}
	\label{fig:additional_vis}
\end{figure*}

\subsubsection{Contributions of Main Ideas} \label{section:main_contributions}
Table \ref{table:main_ablation} highlights the contributions of three key concepts: 1) recurrent temporal fusion, 2) motion-guided BEV fusion, and 3) heatmap-based consistency loss to overall performance. First, Method (a) is obtained by applying recurrent temporal fusion to Baseline-S. This  offers an 8.5\% NDS gain over Baseline-S. 
Adding motion-guided BEV fusion to Method (a) results in Method (b), which further improves the NDS by 1\%. Finally, enabling heatmap-based consistency loss in Method (c) leads to an additional 0.5\% performance gain. Altogether, these three approaches yield a total NDS improvement of 10\%. Notice that the complexity overhead incurred by our method over Baseline-S is only $(205.7-192.2)/205.7=6.56$\%. Since Heatmap-based Temporal Consistency operates only during the training phase, it does not increase the complexity in run-time.

We also compare our method with the parallel fusion approach, \textbf{Baseline-M}. The performance of parallel fusion becomes comparable to that of recurrent fusion only when 17 frames are aggregated. In all other cases, recurrent fusion significantly outperforms parallel fusion.




\subsubsection{Effect of Motion-Guidance in MBFNet}

Table \ref{tab:alignment_ablation} presents the impact of incorporating motion features for BEV feature alignment on overall performance. The baseline model is derived by disabling MFE in Method (b) of Table \ref{table:perception_ablation}. Here, ``w diff." refers to differential encoding without channel-wise attention, while ``w diff. + CWA" indicates the use of both differential encoding and channel-wise attention. Incorporating differential encoding alone improves mAP by 0.3\% and NDS by 0.2\% over the baseline, and adding CWA provides an additional gain of 0.4\% in mAP and 0.3\% in NDS. These results demonstrate that motion features significantly enhance feature alignment.




\begin{figure*}[t]
	\centering
        \centerline{\includegraphics[width=0.80\linewidth]{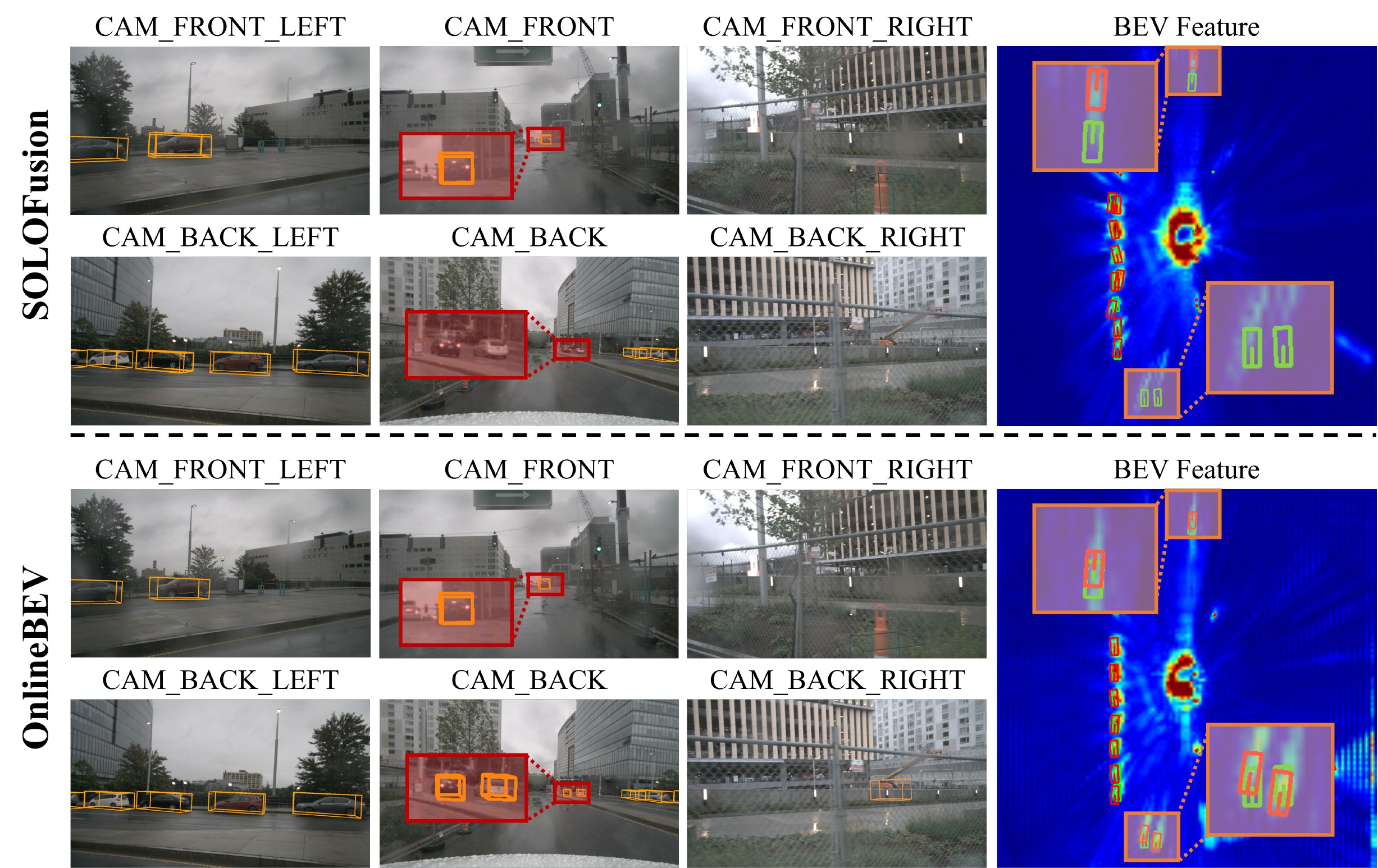}}
        \caption {{\bf Qualitative results of SOLOFusion and OnlineBEV.} Predicted 3D boxes are shown in red, while ground-truth 3D boxes are shown in green.}
	\label{fig:qualitative_results}
\end{figure*}
\input{table/computation_main}

\subsubsection{Ablation Study Conducted on Argoverse 2 Valid Set} \label{ablation:agoverse}

Table \ref{table:agoverse} compares the performance of OnlineBEV against Baseline-S and Baseline-M (17) on the Argoverse 2 \textit{validation} set, evaluating the generalization capability of our method beyond the nuScenes dataset. OnlineBEV demonstrates superior performance, surpassing Baseline-M (17) by 0.9\% in mAP and 1.9\% in CDS.

\subsubsection{Robustness to Corrupted Image Input}

Table \ref{table:data_corruption} highlights the robustness of OnlineBEV under adverse real-world conditions. As illustrated in Fig. \ref{fig:data_corruption}, we applied synthetic input corruptions to simulate motion blur and occlusion scenarios, following the protocols of nuScenes-C \cite{nuscenesC} and nuScenes-R \cite{nuscenesR}. Across all corruption types, OnlineBEV consistently outperforms SOLOFusion \cite{solofusion}. Under the clean setting, OnlineBEV achieves improvements of 1.7\% in mAP and 1.1\% in NDS over SOLOFusion. Notably, the gaps further widen under adverse conditions, reaching 7.4\% mAP and 4.6\% NDS under motion blur, and 5.9\% in mAP and 6.5\% in NDS under occlusion. These results clearly demonstrate the superior robustness of OnlineBEV in handling challenging visual conditions.

\subsubsection{Complexity Analysis}
Table~\ref{table:computation_rev_mAP} summarizes the computational cost and memory usage of OnlineBEV compared to other methods, including SOLOFusion~\cite{solofusion}, StreamPETR~\cite{streampetr}, and SparseBEV~\cite{sparsebev}. All measurements were conducted on an NVIDIA RTX 3090 with a ResNet-50 backbone. BEV-based approaches such as OnlineBEV and SOLOFusion incur higher computational costs than sparse query-based methods like StreamPETR and SparseBEV, as they operate on dense BEV representations. However, they offer global scene understanding and support a wider range of 3D perception tasks, including segmentation and occupancy prediction (see Section~\ref{experiments:3d_perception}). Notably, OnlineBEV achieves the highest accuracy (44.4\% mAP and 54.5\% NDS) while requiring 205.7 GFLOPs and 79.3 ms of inference time. Despite its dense structure, it consumes only 3.4 GB of memory—lower than the 3.9 GB used by SOLOFusion.

\subsubsection{Visualization of Feature Alignment}
We illustrate the effectiveness of our feature alignment strategy by visualizing BEV features over time. Fig. ~\ref{fig:additional_vis} shows the historical, aligned, and current BEV features at each time step $t$, where $t$ denotes the sequential frame index. To highlight the temporal evolution of alignment, five consecutive frames are visualized. As depicted, moving objects in the BEV space exhibit noticeable spatial misalignment between historical and current features. After applying MBFNet, the aligned features demonstrate substantially improved spatial correspondence with the current BEV features, highlighting the effectiveness of our alignment strategy.

\subsubsection{Qualitative Results}
Fig. ~\ref{fig:qualitative_results} presents qualitative comparisons between OnlineBEV and SOLOFusion~\cite{solofusion}. OnlineBEV achieves higher accuracy in localizing distant objects, benefiting from its temporal alignment strategy. Moreover, while SOLOFusion fails to detect objects on the rear side, OnlineBEV successfully identifies them. These results highlight OnlineBEV’s ability to leverage long-term spatio-temporal information more effectively through temporal feature alignment.

\section{Conclusions}

In this paper, we proposed a novel temporal fusion framework for multi-camera 3D perception. Conventional temporal fusion methods often suffer from misaligned BEV features across consecutive frames due to the motion of dynamic objects, which limits their performance. To address this challenge, we introduced OnlineBEV, an effective feature alignment technique tailored for recurrent temporal fusion. OnlineBEV continuously updates historical BEV features by spatially aligning them and fusing them with current BEV features. Our approach leverages spatio-temporal deformable attention to adaptively align BEV features across frames, guided by motion context features extracted from two adjacent frames. To further enhance alignment, we incorporated temporal consistency learning, providing explicit supervision for BEV feature alignment.

Experiments on the nuScenes public dataset demonstrated that OnlineBEV achieves significant performance gains over existing temporal fusion methods. Moreover, when combined with recurrent temporal fusion, our method offers greater computational efficiency compared to parallel fusion approaches.

In this study, we utilized implicit motion features to predict alignment offsets. As a future direction, explicit motion information such as velocity vectors could be extracted from temporal features and integrated into our alignment process. This could be supervised using ground-truth motion data obtained from auxiliary sensors (e.g., LiDAR or radar). We plan to investigate this strategy to further enhance the performance of temporal fusion.

\section{Acknowledgements}
This work was partly supported by 1) Institute of Information \& communications Technology Planning \& Evaluation (IITP) grant funded by the Korea government(MSIT) [NO.RS-2021-II211343, Artificial Intelligence Graduate School Program (Seoul National University)] and 2) the National Research Foundation (NRF) funded by the Korean government (MSIT) (No. RS-2024-00421129).



\bibliographystyle{IEEEtran}
\bibliography{IEEEtran}

\newpage
\section*{Biography Section}
\vspace{-0.8cm}
\begin{IEEEbiography}[{\includegraphics[width=1in,height=1.25in,clip,keepaspectratio]{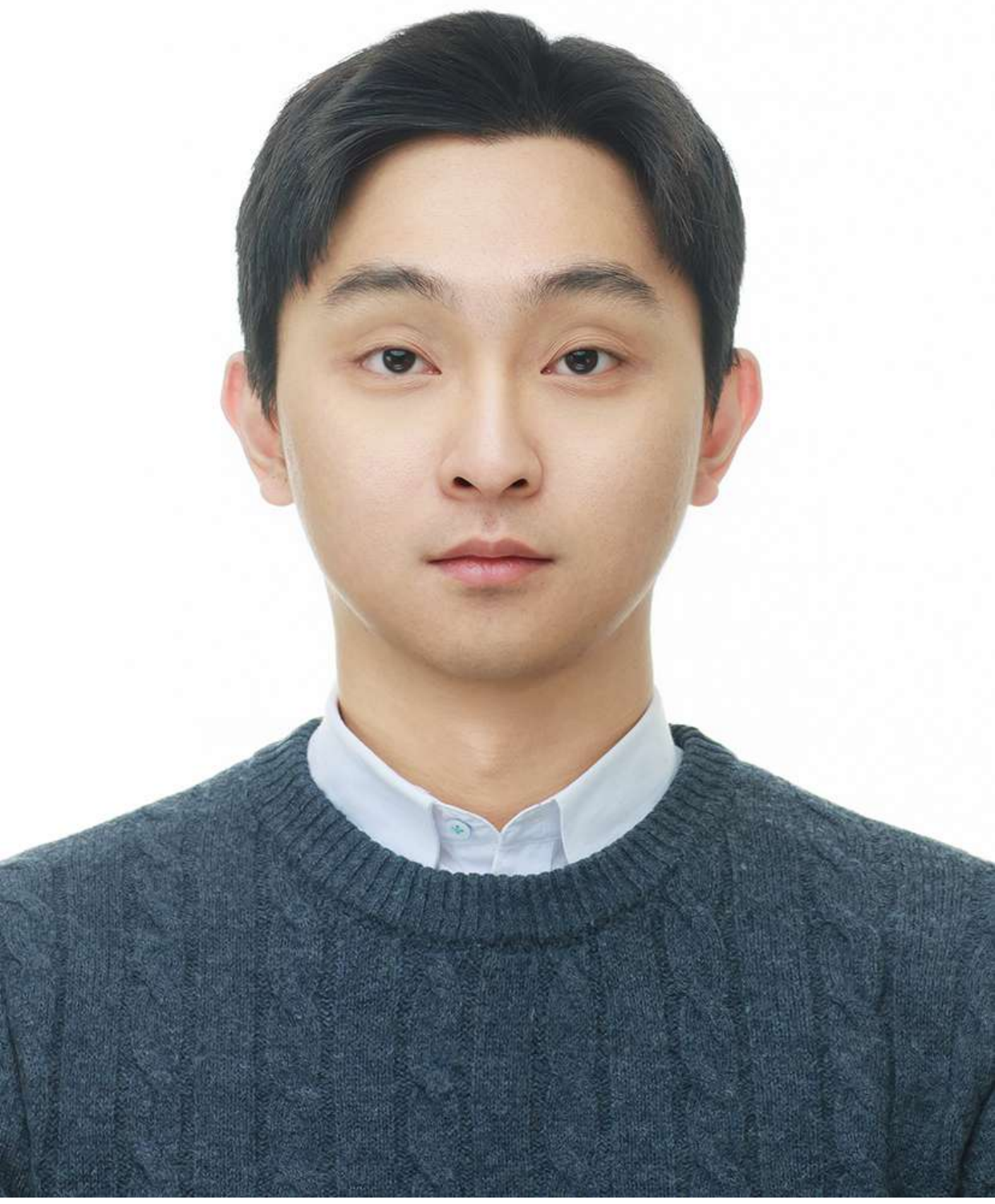}}]{Junho Koh} received his B.S., M.S., and Ph.D. degrees in Electrical Engineering from Hanyang University, Seoul, Korea. In 2024, he joined Hyundai Motor Company, where he has been developing deployable 3D perception systems for end-to-end autonomous driving. His research interests include machine learning, robot vision, and 3D perception for autonomous vehicles.
\end{IEEEbiography}
\vspace{-0.8cm}
\begin{IEEEbiography}[{\includegraphics[width=1in,height=1.25in,clip,keepaspectratio]{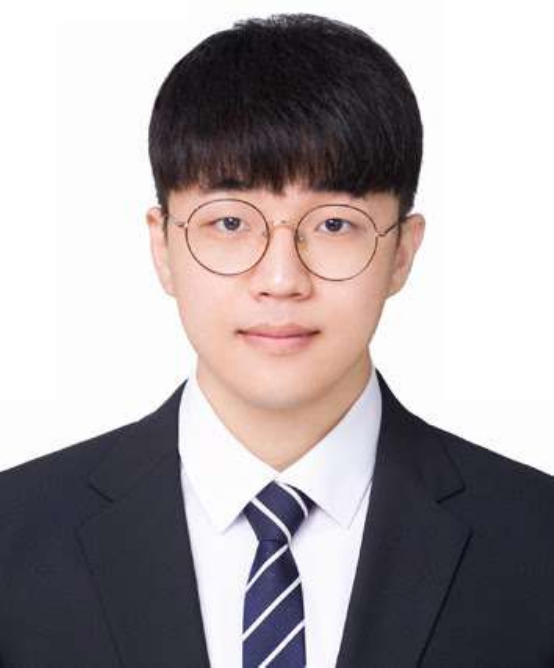}}]{Youngwoo Lee} received his B.S. and M.S. degrees in Electrical Engineering from Hanyang University, Seoul, Korea. In 2024, he joined Samsung Electronics Company, where he has been developing multi-image fusion model for high-resolution image generation. His research interests include super resolution, denoising and tone mapping for software-based image signal processing.
\end{IEEEbiography}
\vspace{-0.8cm}
\begin{IEEEbiography}[{\includegraphics[width=1in,height=1.25in,clip,keepaspectratio]{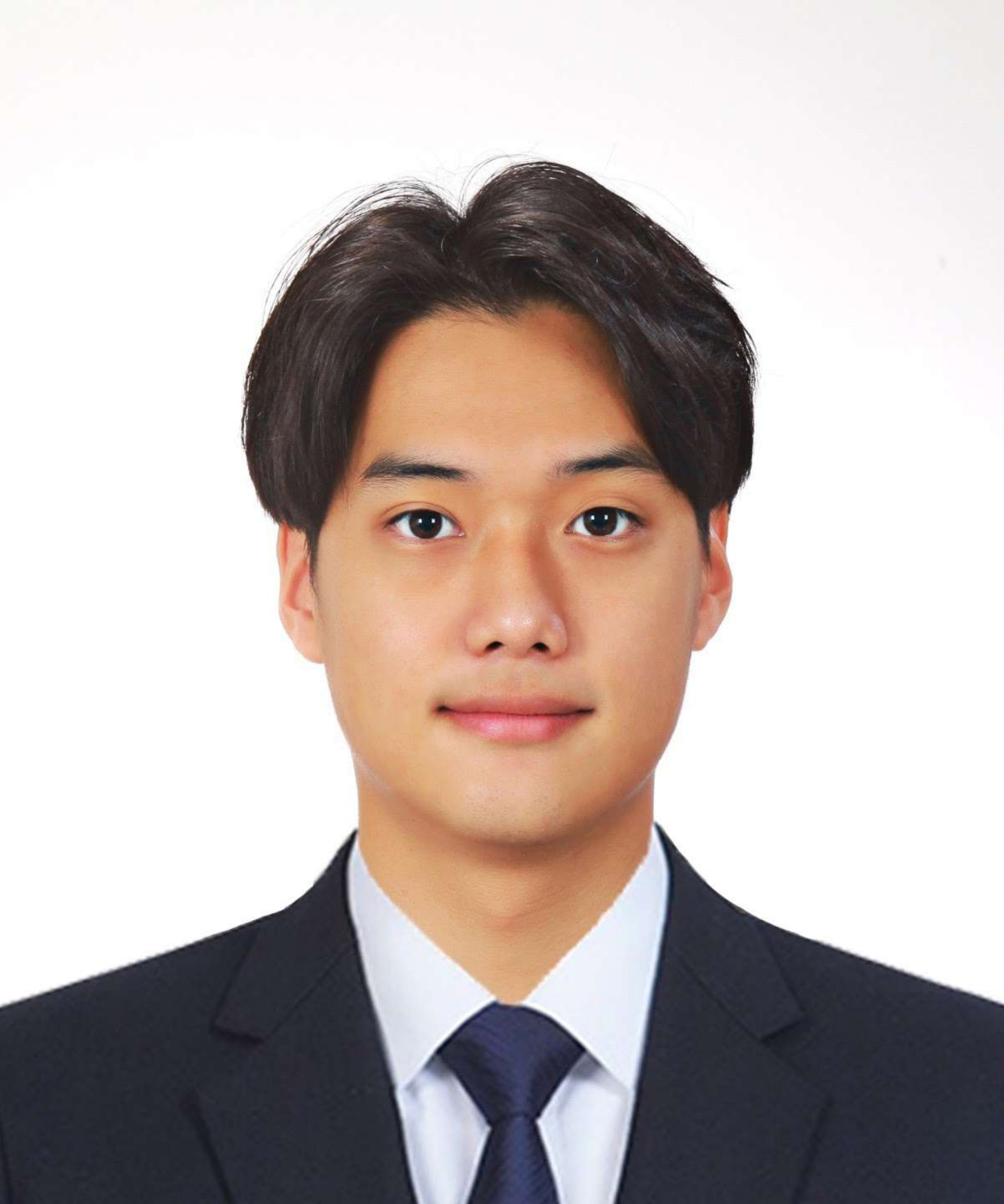}}]{Jungho Kim} received the B.S. degree in Automotive Engineering from Kookmin University, Seoul, Korea, in 2022, and the M.S. degree in Artificial Intelligence from Hanyang University, Seoul, Korea, in 2025. He is currently pursuing a Ph.D. in the Interdisciplinary Program in Artificial Intelligence at Seoul National University. His research interests include 3D perception, trajectory prediction, and planning in autonomous driving.
\end{IEEEbiography}
\vspace{-0.8cm}
\begin{IEEEbiography}[{\includegraphics[width=1in,height=1.25in,clip,keepaspectratio]{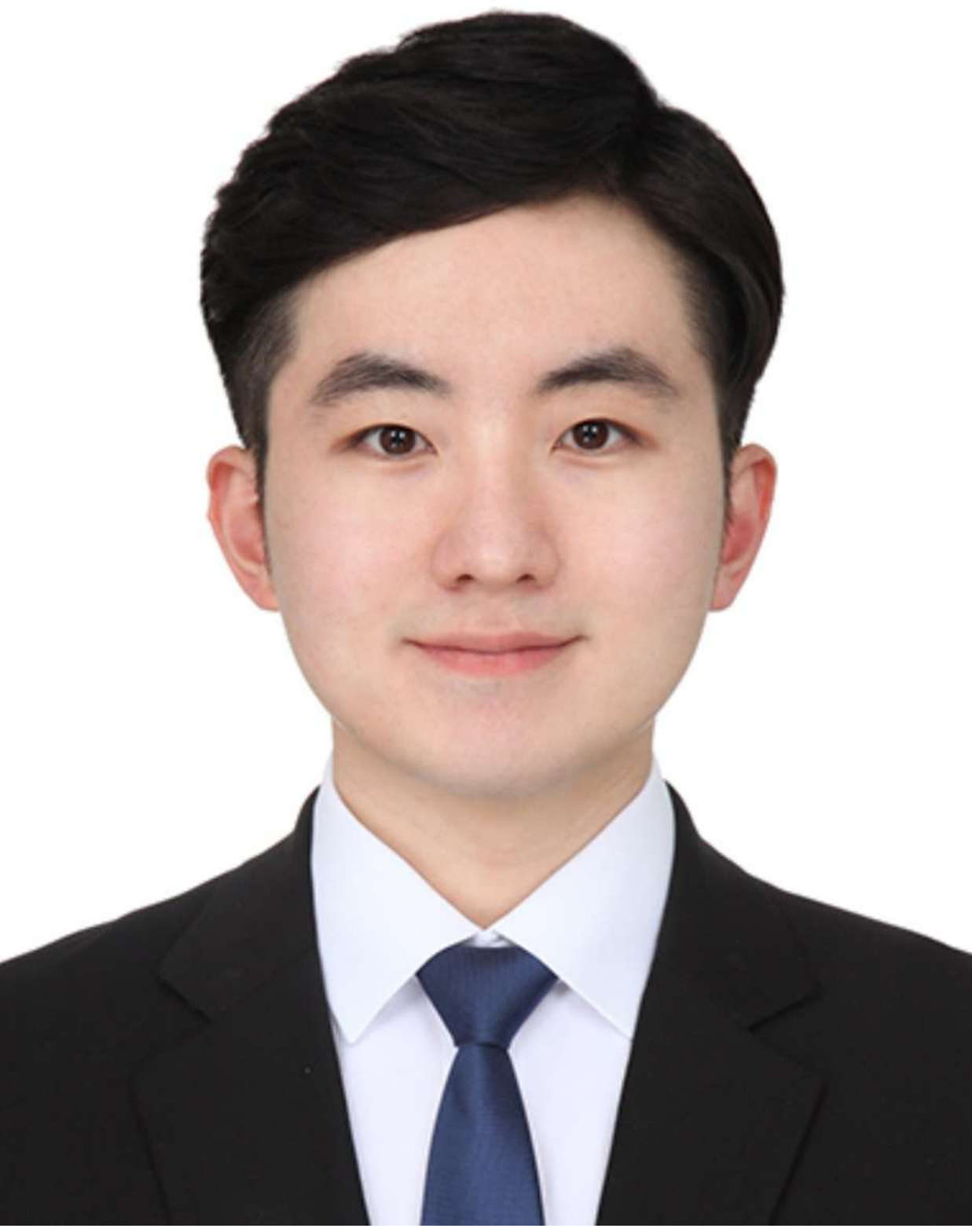}}]{Dongyoung Lee} received the B.S. degree in Electronic Engineering from Kookmin University, Seoul, South Korea, in 2021, and M.S. degree in Artificial Intelligence from Hanyang University, Seoul, Korea, in 2025. He is currently pursuing a Ph.D. in the Department of Electrical and Computer Engineering at Seoul National University. His research interests include deep learning for perception, 3D computer vision, and sensor fusion in autonomous driving.
\end{IEEEbiography}
\vspace{-0.8cm}
\begin{IEEEbiography}[{\includegraphics[width=1in,height=1.25in,clip,keepaspectratio]{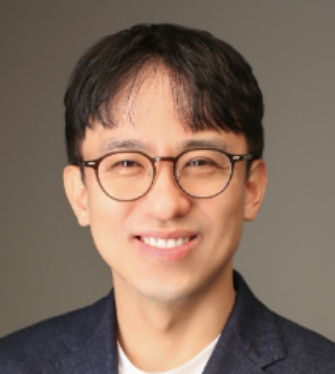}}]{Jun Won Choi} earned his B.S. and M.S. degrees from Seoul National University and his Ph.D. from the University of Illinois at Urbana-Champaign. Following his studies, he joined Qualcomm in San Diego, USA, in 2010. From 2013 to 2024, he served as a faculty member in the Department of Electrical Engineering at Hanyang University. Since 2024, he has held a faculty position in the Department of Electrical and Computer Engineering at Seoul National University. He currently serves as an Associate Editor for both IEEE Transactions on Intelligent Transportation Systems, IEEE Transactions on Vehicular Technology, International Journal of Automotive Technology. His research spans diverse areas including signal processing, machine learning, robot perception, autonomous driving, and intelligent vehicles.
\end{IEEEbiography}
\vspace{-0.8cm}
\vfill

\end{document}

%% file: table/sota_val.tex
\newcolumntype{C}{>{\centering\arraybackslash}p{4.0em}}
\newcolumntype{E}{>{\centering\arraybackslash}p{6.0em}}
\newcolumntype{D}{>{\centering\arraybackslash}p{4.5em}}
\newcolumntype{W}{>{\centering\arraybackslash}p{8.5em}}
\newcolumntype{O}{>{\centering\arraybackslash}p{6.5em}}
\newcolumntype{G}{>{\centering\arraybackslash}p{4.0em}}
\newcolumntype{H}{>{\centering\arraybackslash}p{12.0em}}
\newcolumntype{F}{>{\centering\arraybackslash}p{7.5em}}
\newcolumntype{I}{>{\centering\arraybackslash}p{5.0em}}

\newcolumntype{B}{>{\centering\arraybackslash}p{9.0em}}
\newcolumntype{M}{>{\centering\arraybackslash}p{6.0em}}
\newcolumntype{A}{>{\centering\arraybackslash}p{3.75em}}

\renewcommand{\arraystretch}{1.0}

\begin{table*}[t]
\begin{center}
\caption{{Performance comparison on nuScenes {\it valid} set}}
\begin{adjustbox}{width=0.99\linewidth}

\label{tab:sota_val}
\begin{tabular}{c || E E D | C  C |  C  C  C  C  C}
\Xhline{4\arrayrulewidth}

\textit{Methods} & Backbone & Input Size & \# Frames & mAP $\uparrow$ & NDS $\uparrow$ & mATE $\downarrow$ & mASE $\downarrow$
& mAOE $\downarrow$ & mAVE $\downarrow$ & mAAE $\downarrow$ \\ \hline\hline
PETRv2\cite{petrv2} & ResNet50 & $320 \times 800$ & 2 & 35.0 & 45.6 & 0.726 & 0.277 & 0.505 & 0.503 & 0.181 \\
BEVDepth\cite{bevdepth} & ResNet50 & $256 \times 704$ & 2 & 33.3 & 44.1 & 0.683 & 0.276 & 0.545 & 0.526 & 0.226 \\
BEVStereo\cite{bevstereo} & ResNet50 & $256 \times 704$ & 2 & 34.4 & 44.9 & 0.659 & 0.276 & 0.579 & 0.503 & 0.216 \\
AEDet\cite{aedet} & ResNet50 & $256 \times 704$ & 2 & 35.9 & 47.3 & 0.649 & 0.277 & 0.496 & 0.432 & 0.216 \\
FB-BEV\cite{fbbev} $\dagger$ & ResNet50 & $256 \times 704$ & 2 & 37.8 & 49.8 & 0.620 & 0.273 & 0.444 & 0.374 & 0.200 \\
P2D\cite{p2d} & ResNet50 & $256 \times 704$ & 4 & 37.4 & 48.6 & 0.631 & 0.272 & 0.508 & 0.384 & 0.212 \\
SOLOFusion\cite{solofusion}  & ResNet50 & $256 \times 704$ & 17 & 42.7 & 53.4 & \textbf{0.567} & 0.274 & 0.511 & 0.252 & 0.188 \\
StreamPETR\cite{streampetr}  & ResNet50 & $256 \times 704$ & {\it rnt} & 43.2 & 54.0 & 0.581 & \textbf{0.272} & 0.413 & 0.295 & 0.195 \\
SparseBEV\cite{sparsebev}  & ResNet50 & $256 \times 704$ & 8 & 43.2 & 54.5 & 0.606 & 0.274 & \textbf{0.387} & 0.251 & \textbf{0.186} \\
\rowcolor[gray]{.80}
\textbf{OnlineBEV}  & ResNet50 & $256 \times 704$ & {\it rnt} & \textbf{44.4} & \textbf{54.5} & 0.590 & 0.275 & 0.466 & \textbf{0.244} & 0.197 \\ \hline

StreamPETR\cite{streampetr} $\dagger$ & ResNet50 & $256 \times 704$ & {\it rnt} & 45.0 & 55.0 & 0.613 & \textbf{0.267} & 0.413 & 0.265 & 0.196 \\
SparseBEV\cite{sparsebev} $\dagger$ & ResNet50 & $256 \times 704$ & 8 & 44.8 & 55.8 & 0.581 & 0.271 & \textbf{0.373} & 0.247 & \textbf{0.190} \\
\rowcolor[gray]{.80}
\textbf{OnlineBEV}  $\dagger$ & ResNet50 & $256 \times 704$ & {\it rnt} & \textbf{46.3} & \textbf{56.0} & \textbf{0.552} & 0.271 & 0.450 & \textbf{0.245} & 0.197 \\ \hline

SOLOFusion\cite{solofusion}  & ResNet101 & $512 \times 1408$ & 17 & 48.3 & 58.2 & 0.503 & 0.264 & 0.381 & 0.246 & 0.207 \\
StreamPETR\cite{streampetr} $\dagger$ & ResNet101 & $512 \times 1408$ & {\it rnt} & 50.4 & 59.2 & 0.569 & 0.262 & \textbf{0.315} & 0.257 & 0.199 \\
SparseBEV\cite{sparsebev} $\dagger$ & ResNet101 & $512 \times 1408$ & 8 & 50.1 & 59.2 & 0.562 & 0.265 & 0.321 & 0.243 & \textbf{0.195} \\
\rowcolor[gray]{.80}
\textbf{OnlineBEV}  $\dagger$ & ResNet101 & $512 \times 1408$ & {\it rnt} & \textbf{50.8} & \textbf{59.9} & \textbf{0.508} & \textbf{0.262} & 0.353 & \textbf{0.223} & 0.205 \\ \hline


\Xhline{4\arrayrulewidth}

\end{tabular}
\end{adjustbox}
\end{center}

\end{table*}
\renewcommand{\arraystretch}{1}


%% file: table/sota_test.tex
\renewcommand{\arraystretch}{1.0}

\begin{table*}[t]
\begin{center}
\caption{{ Performance comparison on nuScenes {\it test} set}}

\begin{adjustbox}{width=0.99\linewidth}

\label{tab:sota_test}
\begin{tabular}{c || E E D | C  C |  C  C  C  C  C}
\Xhline{4\arrayrulewidth}

\textit{Methods} & Backbone & Input Size & \# Frames & mAP $\uparrow$ & NDS $\uparrow$ & mATE $\downarrow$ & mASE $\downarrow$
& mAOE $\downarrow$ & mAVE $\downarrow$ & mAAE $\downarrow$ \\ \hline\hline
BEVDepth\cite{bevdepth} & V2-99 & $640 \times 1600$ & 2 & 50.3 & 60.0 & 0.445 & 0.245 & 0.378 & 0.320 & 0.126 \\
BEVStereo\cite{bevstereo} & V2-99 & $640 \times 1600$ & 2 & 52.5 & 61.0 & 0.431 & 0.246 & 0.358 & 0.357 & 0.137 \\
AEDet\cite{aedet} & ConvNeXt-B & $640 \times 1600$ & 2 & 53.1 & 62.0 & 0.439 & 0.247 & 0.344 & 0.292 & 0.130 \\
FB-BEV\cite{fbbev} & V2-99 & $640 \times 1600$ & 2 & 53.7 & 62.4 & 0.439 & 0.250 & 0.358 & 0.270 & 0.128 \\
SOLOFusion\cite{solofusion} & ConvNeXt-B & $640 \times 1600$ & 17 & 54.0 & 61.9 & 0.453 & 0.257 & 0.376 & 0.276 & 0.148 \\
StreamPETR\cite{streampetr} & V2-99 & $640 \times 1600$ & {\it rnt} & 55.0 & 63.6 & 0.479 & \textbf{0.239} & \textbf{0.317} & 0.241 & 0.119 \\
SparseBEV\cite{sparsebev} & V2-99 & $640 \times 1600$ & 8 & 55.6 & 63.6 & 0.485 & 0.244 & 0.332 & 0.246 & \textbf{0.117} \\
\rowcolor[gray]{.80}
\textbf{OnlineBEV} & V2-99 & $640 \times 1600$ & {\it rnt} & \textbf{55.8} & \textbf{63.9} & \textbf{0.424} & 0.253 & 0.362 & \textbf{0.233} & 0.133 \\ \hline


\Xhline{4\arrayrulewidth}

\end{tabular}
\end{adjustbox}
\end{center}

\end{table*}
\renewcommand{\arraystretch}{1}

%% file: table/perception_ablation.tex
\renewcommand{\arraystretch}{1.2}
\begin{table}[t]
\begin{center}
\caption{{ Performance comparison for the other 3D perception tasks on the nuScenes {\it valid} set}}
\begin{adjustbox}{width=0.98\linewidth}
\begin{tabular}{O||W|C||I|I}
\Xhline{4\arrayrulewidth}


\multirow{3}{*}{\begin{tabular}[c]{@{}c@{}} 3D Perception\\Task\end{tabular}} &  \multirow{3}{*}{{\it Method}} & \multirow{3}{*}{{ \# Frames}} &\multicolumn{2}{c}{Performance} \\ \cline{4-5}
 &&&
 \multirow{2}{*}{{mIoU-seg $\uparrow$}} & \multirow{2}{*}{{mIoU-occ $\uparrow$}}  \\ 
 &&& \\

 \hline\hline
\multirow{4}{*}{\begin{tabular}[c]{@{}c@{}} BEV\\Segmentation\end{tabular}} 
& BEVDet-Depth \cite{bevdet} & 1 & 43.6 & - \\ \cline{2-5}
& BEVDepth \cite{bevdepth} & 2 & 44.7 & -\\ \cline{2-5}
& SOLOFusion \cite{solofusion} & 17 & 49.5 & -\\ \cline{2-5} 
 & OnlineBEV & {\it rnt} & 52.4 & -\\
\hline 

\multirow{4}{*}{\begin{tabular}[c]{@{}c@{}} 3D Occupancy\\ Prediction\end{tabular}} 
& BEVDet-Depth \cite{bevdet} & 1 & -  & 40.8 \\ \cline{2-5}
& BEVDepth \cite{bevdepth} & 2 & - & 42.6 
\\ \cline{2-5}
& SOLOFusion \cite{solofusion} & 17 & - & 44.7 \\ \cline{2-5}
 & OnlineBEV & {\it rnt} & - & 45.9 \\

\Xhline{4\arrayrulewidth}
\end{tabular}
\label{table:perception_ablation}
\end{adjustbox}
\end{center}
\end{table}
\renewcommand{\arraystretch}{1}

%% file: table/main_ablation.tex
\renewcommand{\arraystretch}{1.2}
\begin{table*}[t]
\begin{center}
\caption{{ Ablation study to evaluate the contributions of  the individual components}}
\begin{adjustbox}{width=0.98\textwidth}
\label{table:main_ablation}
\begin{tabular}{M|A||F|F|F|B||A|A|A||A|A}
\Xhline{4\arrayrulewidth}
 \multirow{3}{*}{{\it Method}} & \multirow{3}{*}{\# Frames}& \multicolumn{4}{c||}{{\it BEV Temporal Fusion Strategy}} &\multicolumn{3}{c||}{Performance} & \multirow{3}{*}{GFLOPs} & \multirow{3}{*}{FPS} \\ \cline{3-9}
 & & \multirow{2}{*}{\begin{tabular}[c]{@{}c@{}}Parallel \\ Temporal Fusion \end{tabular}}&
 \multirow{2}{*}{\begin{tabular}[c]{@{}c@{}}Recurrent \\ Temporal Fusion \end{tabular}}&
 \multirow{2}{*}{\begin{tabular}[c]{@{}c@{}}Motion-Guided\\BEV Fusion\end{tabular}}& 
 \multirow{2}{*}{\begin{tabular}[c]{@{}c@{}}Heatmap-based\\Temporal Consistency\end{tabular}} &
 \multirow{2}{*}{mAP $\uparrow$} & \multirow{2}{*}{NDS $\uparrow$} & \multirow{2}{*}{mAVE $\downarrow$} &\\ 
 &&&&&&&&&\\
 \hline\hline
 Baseline-S & 1 & & & & & 35.4 & 41.9 & 0.758 & 192.2 & 13.8 \\
\hline 

\multirow{4}{*}{\begin{tabular}[c]{@{}c@{}}
Baseline-M\end{tabular}} & 2 & \checkmark & & & & 37.0 & 46.9 & 0.430 & 194.2 & 13.7 \\
& 5 & \checkmark & & & & 38.9 & 49.2 & 0.310 & 194.8 & 13.6 \\ 
& 17 & \checkmark & & & & 40.8 & 50.7 & 0.287 & 198.7 & 13.4  \\ \cline{2-11}
& 2 & \checkmark & & \checkmark & & 37.9 & 47.3 & 0.401 & 205.7 & 12.6 \\ 
& 5 & \checkmark & & \checkmark & & 40.9 & 50.1 & 0.301 & 249.0 & 7.7 \\ \hline
Method (a) & {\it rnt} & & \checkmark & & & 40.8 & 50.4 & 0.296 & 192.6 & 13.8 \\
Method (b) & {\it rnt} & & \checkmark & \checkmark & & 42.1 & 51.4 & 0.287 & 205.7 & 12.6 \\
Method (c) & {\it rnt} & & \checkmark & \checkmark & \checkmark & 42.5 & 51.9 & 0.280 & 205.7 & 12.6 \\

\Xhline{4\arrayrulewidth}
\end{tabular}
\end{adjustbox}
\end{center}
\end{table*}
\renewcommand{\arraystretch}{1}

%% file: table/alignment_ablation.tex
\renewcommand{\arraystretch}{1.2}
\begin{table}[t]
\begin{center}
\caption{{Ablation study for evaluating the effect of motion guidance of MBFNet}}
\begin{adjustbox}{width=0.98\linewidth}
\label{tab:alignment_ablation}
\begin{tabular}{M||I||A|A|A}

\Xhline{4\arrayrulewidth}


  \multirow{2}{*}{{\it MFE Block}} &\multicolumn{4}{c}{Performance} \\ \cline{2-5}
 &
 mAP $\uparrow$ & NDS $\uparrow$ & mATE $\downarrow$ & mAVE $\downarrow$ \\ 

 \hline\hline

 Baseline & 41.4 & 50.9 & 0.614 & 0.295 \\ \cline{1-5}
  w diff.   & 41.7 & 51.1 & 0.613 & 0.289  \\
  w diff. + CWA & 42.1 & 51.4 & 0.611 & 0.287  \\

\Xhline{4\arrayrulewidth}
\end{tabular}
\end{adjustbox}
\end{center}
\end{table}
\renewcommand{\arraystretch}{1}

%% file: table/agoverse.tex
\newcolumntype{X}{>{\centering\arraybackslash}p{6.7em}}
\newcolumntype{Y}{>{\centering\arraybackslash}p{3.6em}}

\renewcommand{\arraystretch}{1.2}
\begin{table}[t]
\begin{center}
\caption{ Ablation study conducted on Agoverse 2 dataset}
\begin{adjustbox}{width=0.99\linewidth}
\label{table:agoverse}
\begin{tabular}{X||Y Y|Y Y Y}
\Xhline{4\arrayrulewidth}
Method & mAP $\uparrow$ & CDS $\uparrow$ & mATE $\downarrow$ & mASE $\downarrow$ & mAOE $\downarrow$ \\ \hline
Baseline-S & 13.6 & 8.2 & 0.967 & 0.452 & 0.854 \\ \hline
Baseline-M (17) & 14.7 & 10.4 & 0.938 & 0.431 & 0.782 \\ \hline
OnlineBEV & 15.6 & 12.3 & 0.899 & 0.411 & 0.746 \\ \hline


\Xhline{4\arrayrulewidth}
\end{tabular}
\end{adjustbox}
\end{center}
\end{table}
\renewcommand{\arraystretch}{1}

%% file: table/data_corruption_ablation.tex
\newcolumntype{Q}{>{\centering\arraybackslash}p{5.3em}}

\renewcommand{\arraystretch}{1.2}
\begin{table}[t]
\begin{center}
\caption{ Performance comparison on the corrupted nuScenes test set}
\begin{adjustbox}{width=0.95\linewidth}
\label{table:data_corruption}
\begin{tabular}{Q||Q|Q|Q|Q}
\Xhline{5\arrayrulewidth}

\multirow{2}{*}{{\textit{Test Input}}} & \multicolumn{2}{c|}{SOLOFusion} & \multicolumn{2}{c}{OnlineBEV} \\ \cline{2-3} \cline{4-5} 
 & mAP $\uparrow$ & NDS $\uparrow$ & mAP $\uparrow$ & NDS $\uparrow$ \\  \hline
 Original & 42.7 & 53.4 & 44.4 & 54.5 \\  \hline
 Motion Blur & 32.2 ($\downarrow$ 10.5) & 46.9 ($\downarrow$ 6.5) & 39.6 ($\downarrow$ 4.8) & 51.5 ($\downarrow$ 3.0) \\ \hline
 Occlusion & 29.7 ($\downarrow$ 13.0) & 42.3 ($\downarrow$ 11.1) & 35.6 ($\downarrow$ 8.8) & 48.8 ($\downarrow$ 5.7) \\


\Xhline{5\arrayrulewidth}
\end{tabular}
\end{adjustbox}
\end{center}
\end{table}
\renewcommand{\arraystretch}{1}

%% file: table/computation_main.tex
\newcolumntype{X}{>{\centering\arraybackslash}p{7.0em}}
\newcolumntype{Y}{>{\centering\arraybackslash}p{5.0em}}

\renewcommand{\arraystretch}{1.2}
\begin{table*}[t]
\begin{center}
\caption{ Computational complexity and memory requirements}
\begin{adjustbox}{width=0.9\linewidth}

\begin{tabular}{X|X || Y|Y|Y|Y|Y|Y}
\Xhline{4\arrayrulewidth}

\multicolumn{2}{c||}{\multirow{2}{*}{\it Method}} & 
\multirow{2}{*}{mAP $\uparrow$} & 
\multirow{2}{*}{NDS $\uparrow$} & 
\multirow{2}{*}{\begin{tabular}[c]{@{}c@{}} Memory \\ (GB)\end{tabular}} & 
\multirow{2}{*}{\begin{tabular}[c]{@{}c@{}} Inference \\ Time (ms)\end{tabular}} & 
\multirow{2}{*}{GFLOPs} & 
\multirow{2}{*}{\begin{tabular}[c]{@{}c@{}} Params \\ (M)\end{tabular}} \\
\multicolumn{2}{c||}{} & & & & & & \\ \hline \hline

\multirow{2}{*}{\begin{tabular}[c]{@{}c@{}} Query-based \\ Method\end{tabular}} 
& SparseBEV   & 43.2 & 54.5 & 4.1 & 43.2 & 192.0 & 44.34 \\ \cline{2-8}
& StreamPETR  & 43.2 & 54.0 & 2.3 & 37.5 & 145.7 & 37.21 \\ \hline

\multirow{2}{*}{\begin{tabular}[c]{@{}c@{}} BEV-based \\ Method\end{tabular}} 
& SOLOFusion  & 42.7 & 53.7 & 3.9 & 72.8 & 198.7 & 64.99 \\ \cline{2-8}
& OnlineBEV   & 44.4 & 54.5 & 3.4 & 79.3 & 205.7 & 65.03 \\ \hline

\Xhline{4\arrayrulewidth}
\end{tabular}
\end{adjustbox}
\label{table:computation_rev_mAP}
\end{center}
\end{table*}
\renewcommand{\arraystretch}{1}